\documentclass[11pt, a4paper, logo, twocolumn, external, copyright]{googledeepmind}

\pdftrailerid{redacted}

\makeatletter
\renewcommand\bibentry[1]{\nocite{#1}{\frenchspacing\@nameuse{BR@r@#1\@extra@b@citeb}}}
\makeatother

\usepackage{kantlipsum, lipsum}
\usepackage{dsfont}
\usepackage{gdm-colors}

\usepackage[authoryear, sort&compress, round]{natbib}

\renewcommand{\paragraph}[1]{\noindent\textbf{#1}}

\graphicspath{{figures/}}

\title{Compression Scaling Laws: \\ Unifying Sparsity and Quantization}

\correspondingauthor{evcu@google.com}


\reportnumber{001} 

\author[1,2]{Elias Frantar}
\author[*,1]{Utku Evci}
\author[1]{Wonpyo Park}
\author[1]{Neil Houlsby}
\author[2]{Dan Alistarh}

\affil[*]{Corresponding author}
\affil[1]{Google DeepMind}
\affil[2]{Institute of Science and Technology Austria}

\begin{abstract}

We investigate how different compression techniques---such as weight and activation quantization, and weight sparsity---affect the scaling behavior of large language models (LLMs) during pretraining. Building on previous work showing that weight sparsity acts as a constant multiplier on model size in scaling laws, we demonstrate that this ``effective parameter'' scaling pattern extends to quantization as well. Specifically, we establish that weight-only quantization achieves strong parameter efficiency multipliers, while full quantization of both weights and activations shows diminishing returns at lower bitwidths. Our results suggest that different compression techniques can be unified under a common scaling law framework, enabling principled comparison and combination of these methods.
\end{abstract}

\begin{document}

\maketitle

\newcommand{\expect}[2]{\mathds{E}_{{#1}} \left[ {#2} \right]}
\newcommand{\myvec}[1]{\boldsymbol{#1}}
\newcommand{\myvecsym}[1]{\boldsymbol{#1}}
\newcommand{\vx}{\myvec{x}}
\newcommand{\vy}{\myvec{y}}
\newcommand{\vz}{\myvec{z}}
\newcommand{\vtheta}{\myvecsym{\theta}}

\section{Introduction}

Recent advances in large language models (LLMs) are based on the principle that model performance improves predictably with increased model size and training data, following well-characterized scaling laws~\citep{kaplan2020scalinglawsneurallanguage}. At the same time, the computational and memory costs of  large models have driven significant interest in model compression techniques like quantization and sparsification. While model compression has been studied extensively in isolation, and in particular for post-training compression, we do not yet have a good understanding about how they interact with scaling of LLMs when performing compressed training from scratch. 

\noindent\textbf{The Compression Scaling Law.} Understanding these relationships  not only provides theoretical insights into compression, but also enable more principled approaches to building efficient language models. This report analyzes  how different forms of compression affect model scaling through an ``effective parameter count'' parameter that is associated to each form of compression. 
We show that, across both sparsity and quantization, the following general \textbf{compressed training law} holds:

\begin{equation} 
\label{eqn:main_law}
L(N, D, C) = \frac{a}{(N \cdot \text{eff}(C))^b} + \frac{c}{D^d} + e,
\end{equation}

\noindent where $N$ is the number of parameters, $D$ is the amount of data, $C$ is the compression type, and $N \cdot \text{eff}(C)$ denotes the \textbf{effective parameter count} for a given form of compression $C$, which would simply be $N$ for full precision. 
The form of this law is derived following prior work on sparsity scaling laws~\citep{frantar2023scalinglawssparselyconnectedfoundation}.

Intuitively, $\text{eff}(C)$ indicates, on average, how much information one parameter compressed using the compressed representation $C$ (e.g., sparsity or quantization) is worth relative to a standard uncompressed BF16 parameter. For low compression, we expect this value to be close to 1; for high compression, it should be close to 0, based on the intuition that a 1-bit parameter cannot carry the same amount of information as a 16-bit one. 
Assuming this model is correct, up to constants, there are several notable consequences:

\begin{enumerate}
    \item To determine $\text{eff}(C)$, we only need to sweep over model sizes with fixed data, e.g., Chinchilla-optimal training~\citep{hoffmann2022trainingcomputeoptimallargelanguage}. In particular, we do not have to vary data amounts, leading to faster exploration.
    \item Second, if we determine this multiplier over short training runs (within reason), we  should still be able to transfer results to longer training runs. Notably, in the case of sparsity scaling laws, longer training durations were actually \emph{necesary} to reach  optimality for training over compressed models~\citep{frantar2023scalinglawssparselyconnectedfoundation}.
    \item Finally, a simple implication of this is formula the Chinchilla optimal amount of training data to use for a model compressed using technique $C$ is the Chinchilla-optimal amount of its uncompressed version, of size $N$, multiplied  by $\text{eff}(C)$. In other words, the parameter/size ratio remains the same, but model size is now expressed in terms of effective parameter count.
\end{enumerate}

The rest of this technical report is organized as follows: we first discuss our experimental setup in Section~\ref{sec:experimental_setup}. We then validate and analyze compression scaling laws for weight-only quantization, weight-and-activation quantization, and sparsity, in Section~\ref{sec:scaling_law_weight_quant}. We discuss related and concurrent work in Section~\ref{sec:related-work}.  We conclude in Section~\ref{sec:discussion}.

\section{The Compression Scaling Law}
\label{sec:scaling_law_weight_quant}

\subsection{Proposed Law}

\paragraph{Definition.} We posit that, up to constants, the following general \textbf{compressed scaling law} for training holds:

\begin{equation} 
\label{eqn:main_law}
L(N, D, C) = \frac{a}{(N \cdot \text{eff}(C))^b} + \frac{c}{D^d} + e,
\end{equation}

\noindent where $N$ is the number of parameters, $D$ is the amount of data, $C$ is the compression type, $e$ is the irreducible error, and $N \cdot \text{eff}(C)$ denotes the \textbf{effective parameter count} for a given form of compression $C$, which would simply be $N$ for full precision training without compression. Thus, we call $\text{eff}(C)$ the \textbf{effective parameter multiplier (EPM)} that depends on the model architecture and compression type, but is independent of the parameter count or dataset size. 

The rest of this paper is dedicated to experimental validation of this law, and an exploration of its implications for sparse and quantized representations. 

\subsection{Experimental Setup}
\label{sec:experimental_setup}

\paragraph{Models and data.} 
We begin by detailing the experimental setup for validating the scaling law. 
Our experiments focus on Llama-type models~\citep{touvron2023llama2openfoundation} trained on a variant of the C4 dataset~\citep{raffel2020exploring}, following the Chinchilla ratios for tokens-per-parameter scaling, and well- optimized learning rate and batch size combinations for the setting.

\paragraph{Hyper-parameter transfer.}
One key aspect of scaling studies involving compression is that different hyper-parameter tuning between compressed and uncompressed baseline can skew results in a major way. For instance, training the compressed model with much higher learning rate can be a major advantage~\citep{wang2023bitnet}. At the same time, a poor choice of hyper-parameters can significantly disadvantage compressed training. 

\begin{figure}[h]
\centering
\includegraphics[width=0.45\textwidth]{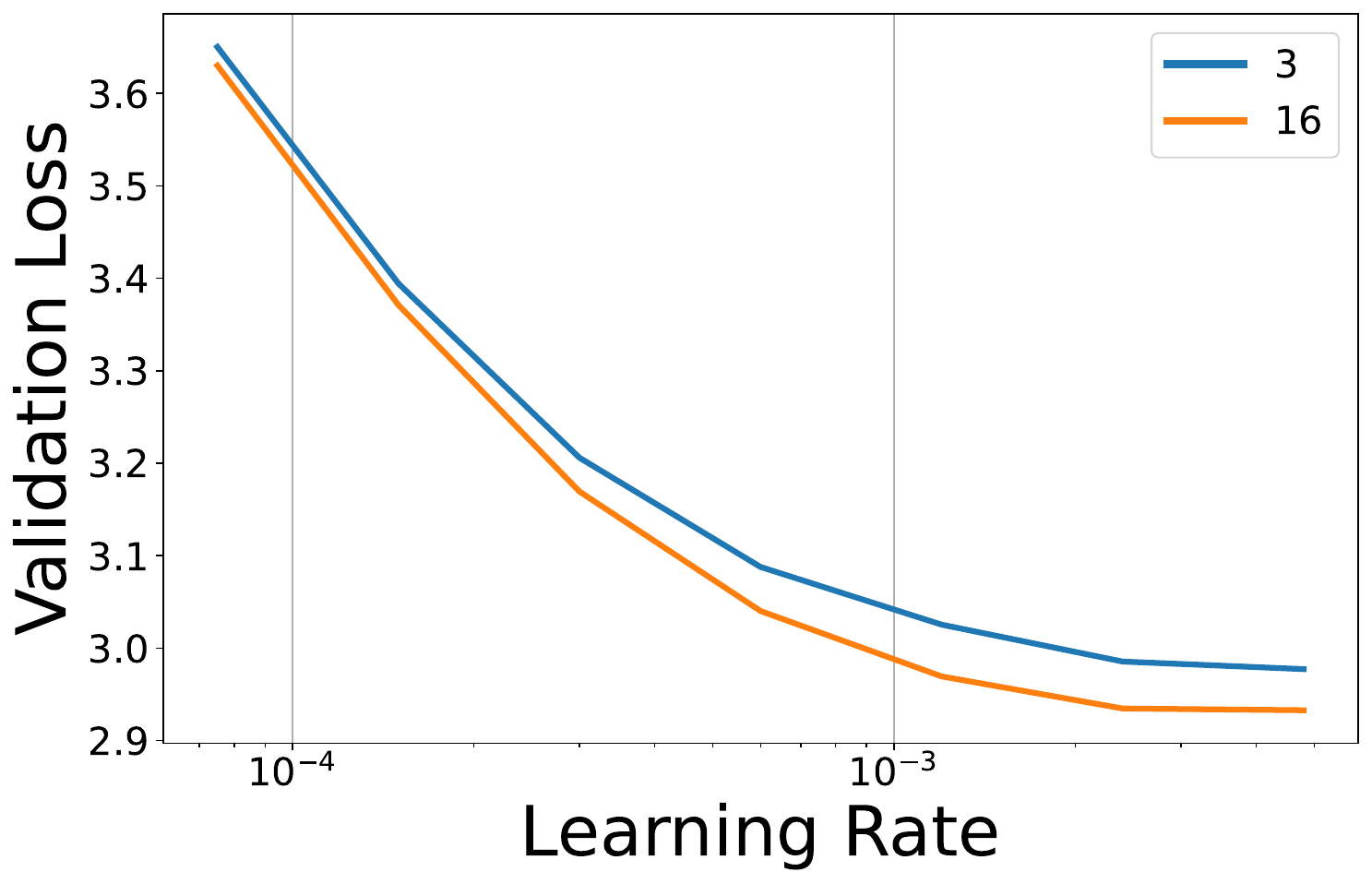}
\caption{Learning rate sweep comparison for BF16 and INT3 models.}
\label{fig:lr_INT3}
\end{figure}

Fortunately, it appears that good hyper-parameter settings for standard dense training seem to transfer quite well also to compressed training. In Figure~\ref{fig:lr_INT3} we illustrate this by showing the results of a learning rate sweep for models trained using BF16 and INT3 weights, finding that the relative ordering of settings remains identical, in both cases.\footnote{We were not able to reproduce some prior claims in literature that compressed models tolerate higher learning rates during training; both models seem to diverge at similar values in our experiments. } 
Thus, we always use the same hyper-parameters for dense and compressed training, guaranteeing fairness relative to the baseline while seemingly not hindering compression accuracy by too much.

\paragraph{Data independence.}
One of our key simplifying assumptions is that compression multipliers are approximately \emph{constant}, in particular, they are independent of the amount of training data used. This has been suggested by~\citet{frantar2023scalinglawssparselyconnectedfoundation}; however, we perform some further validation in our experimental setup. For this, we repeat sweeps with two particularly interesting compression settings at 2x Chinchilla-training and compare the resulting lines to those predicted from the fitted multipliers of our 1x experiments. As can be seen in Figure~\ref{fig:data_independence} below, the lines are virtually identical, thus providing further evidence of the data independence.

\begin{figure}[h]
\centering
\includegraphics[width=0.45\textwidth]{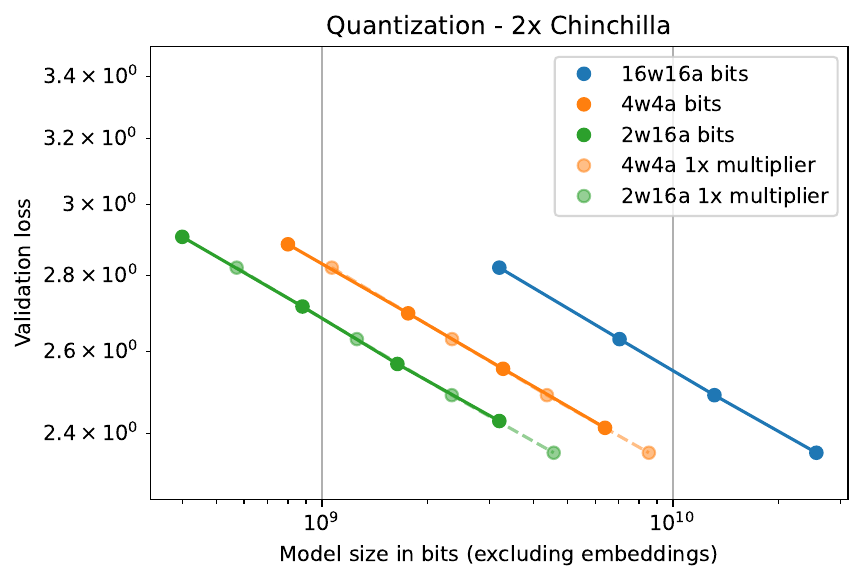}
\caption{Validation of the independence between scaling laws and the amount of data used. The legend represents the number of bits per weight.}
\label{fig:data_independence}
\end{figure}

\begin{table*}[h]
\centering
\begin{tabular}{|l|c|}
\hline
\textbf{Llama 100M for 10B tokens @ 2-bit} & \textbf{Validation loss} \\
\hline
Centered = 2 zero point & 3.268 \\
Symmetric = 1.5 zero point & 3.250 \\
\hline
\end{tabular}
\caption{Tuning results for 2-bit quantization.}
\label{tab:weight-only-2bit}
\end{table*}

\paragraph{Quantized estimation.}
For weight quantization, we roughly follow BitNet~\citep{wang2023bitnet, ma2024era1bitllmslarge} and train quantized models via vanilla Straight-Through Estimation (STE). That is, we run forward passes with quantized weights, with scales computed dynamically from weight statistics. (As opposed to classic methods such as PACT~\citep{choi2018pact} or LSQ~\citep{esser2019learned}, scale parameters are not learned.) 

\subsection{Validation: Weight-Only Quantization}

\begin{figure*}[h]
\centering
\includegraphics[width=0.45\textwidth]{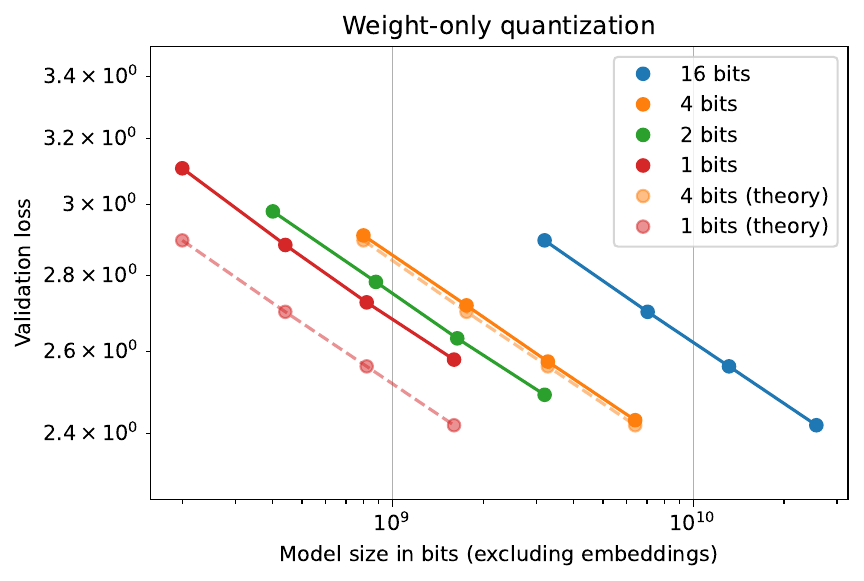}
\includegraphics[width=0.45\textwidth]{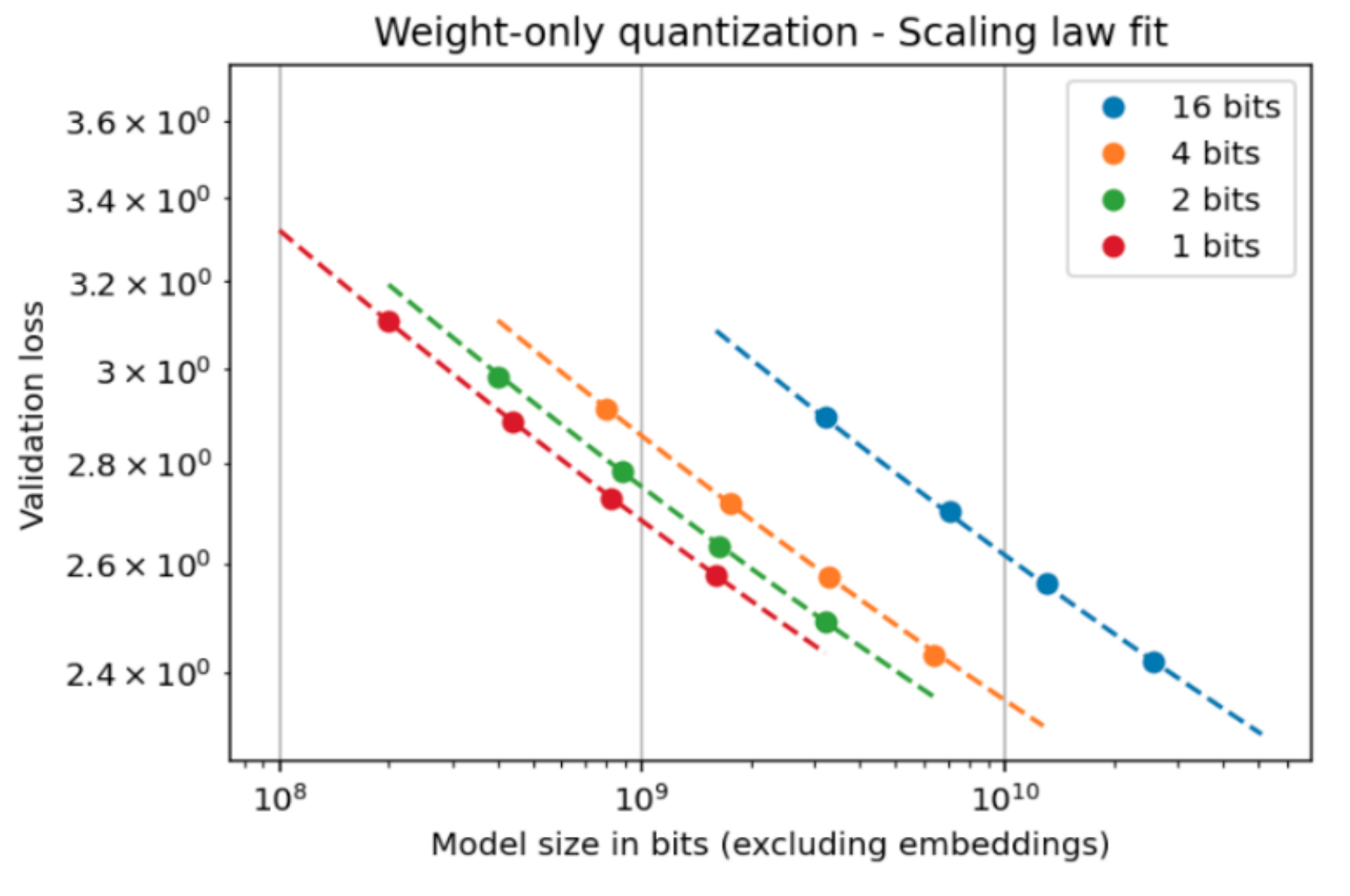}
\caption{Scaling results (loss and fit) for weight-only quantization.}
\label{fig:scaling-weight-quantization}
\end{figure*}

\begin{figure*}[ht!]
\centering
\includegraphics[width=0.45\textwidth]{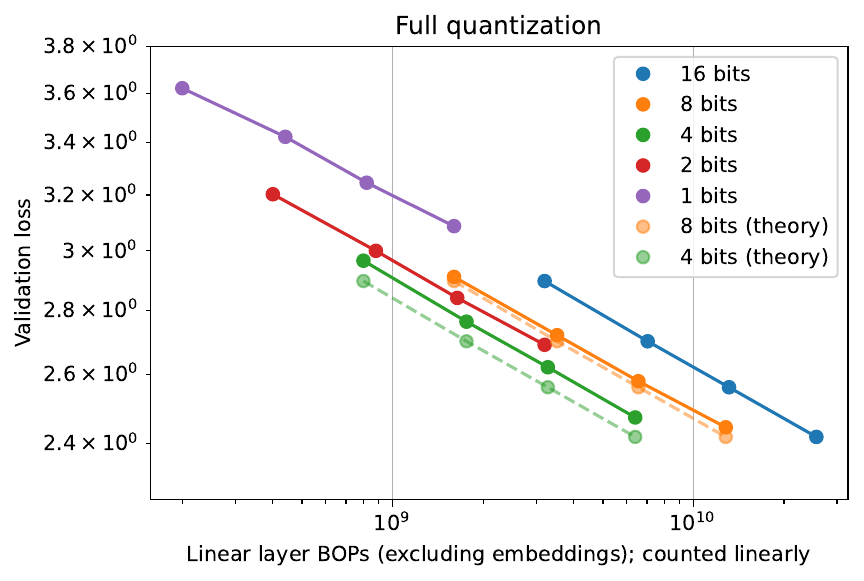}
\includegraphics[width=0.45\textwidth]{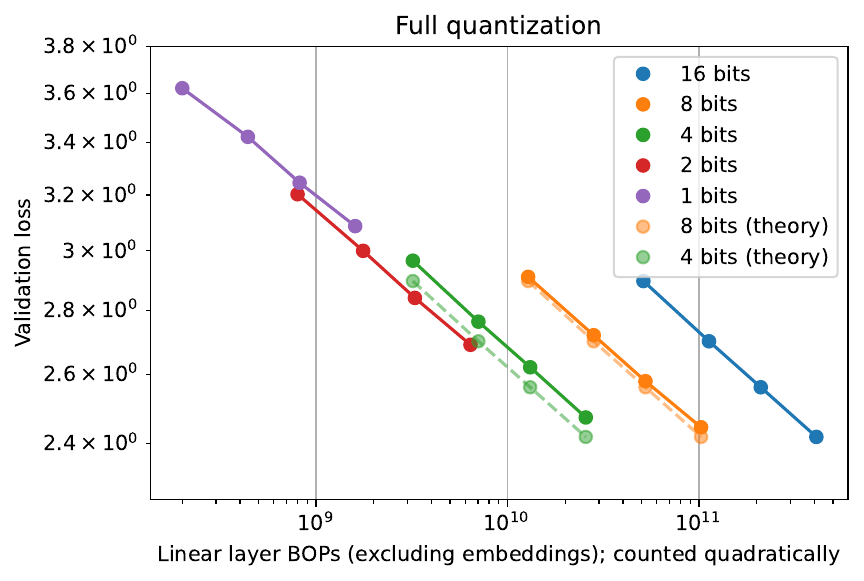}
\caption{Scaling results for full quantization with linear and quadratic speedup counting.}
\label{fig:full-quantization}
\end{figure*}

Scaling experiments with weight-only quantization, shown in Figure~\ref{fig:scaling-weight-quantization} show clear trends, both in terms of raw data, as well as in terms of the scaling law fit. The corresponding bitwidth-effectiveness multipliers, shown in Table~\ref{tab:multipliers-eff-parameters}, suggest that 4-bit is indeed relatively close to lossless (matching state-of-the-art post-training quantization results). Significant size reductions at the same loss level can be made by further dropping the bit-width, coupled with corresponding quantization-aware pretraining. Although returns are clearly diminishing, further improvements still appear quite useful, e.g., 4-bit $\rightarrow$ 1-bit seems to give $\sim$2x improvement.

\begin{table}[h]
\centering
\begin{tabular}{|l|c|c|c|}
\hline
& 4-bit & 2-bit & 1-bit \\
\hline
EPM & 0.923 & 0.702 & 0.466 \\
Size gain & 3.69x & 5.62x & 7.46x \\
\hline
\end{tabular}
\caption{Effective parameter multipliers (EPM) and size gains for different bit-widths.}
\label{tab:multipliers-eff-parameters}
\end{table}

\subsection{Full Weight-and-activation Quantization}

We now extend the above approach to quantize not just the weights but also the corresponding activations. We note that, in these experiments, we only quantize inputs to all linear, non-embedding layers; this means attention itself is currently not quantized.
The results are shown in Figure~\ref{fig:full-quantization}, with two variants: with linear and quadratic speedup counting.

\paragraph{Scaling.} We observe that, using the quantization strategy described in Section~\ref{sec:experimental_setup}, we can train low-bitwidth models in a surprisingly stable manner. One interesting observation is that when counting speedups linearly (i.e., 4-bit being 4x faster, as is the case on current hardware), then bit-widths lower than 4 are clearly not Pareto optimal. However, if we count speedups quadratically (the theoretical scaling of multiplication cost), 2-bit still brings a noticeable improvements. Moreover, the 1-bit option is far from optimal, but it is surprising that it trains stably.

\begin{figure}[t]
\centering
\includegraphics[width=0.45\textwidth]{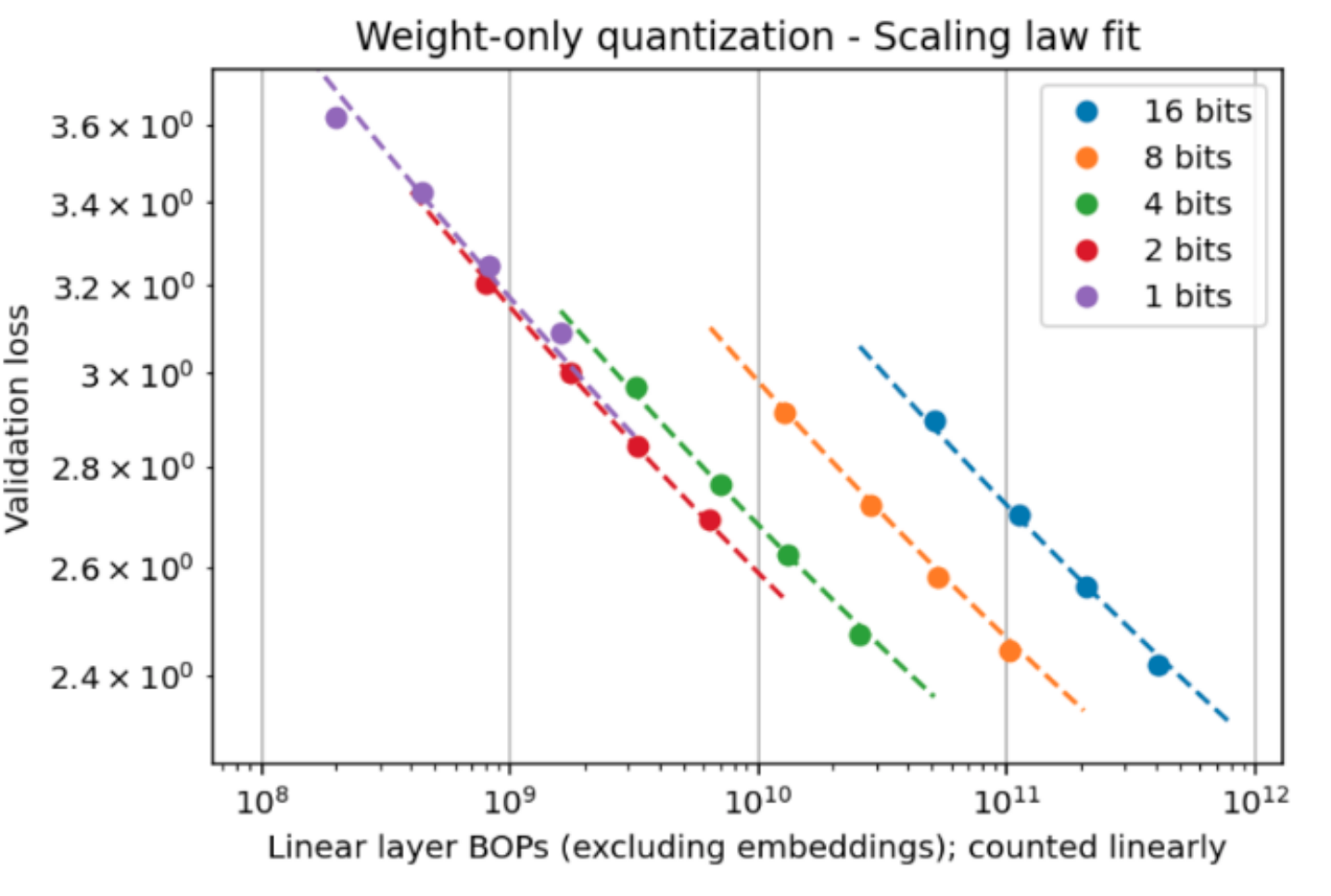}
\caption{Scaling law fit for full quantization.}
\label{fig:full-quantization-fit}
\end{figure}

The fit between the empirical and the predicted law is provided in Figure~\ref{fig:full-quantization-fit}. While this fit is not quite as clean as for weight-only quantization, it still looks very reasonable, especially for more than $ 1$-bit per parameter. Overall, we obtain reasonable  effectiveness multipliers up to 4-bit, after which they drop off quite sharply.

\begin{table}[h]
\centering
\begin{tabular}{|l|c|c|c|c|}
\hline
& 8-bit & 4-bit & 2-bit & 1-bit \\
\hline
EPM & 0.857 & 0.747 & 0.289 & 0.067 \\
\hline
\end{tabular}
\caption{Effective parameter multipliers (EPM) for full quantization.}
\end{table}

\subsection{Mixed Activation-Weight Quantization}

\begin{figure}[h]
\centering
\includegraphics[width=0.45\textwidth]{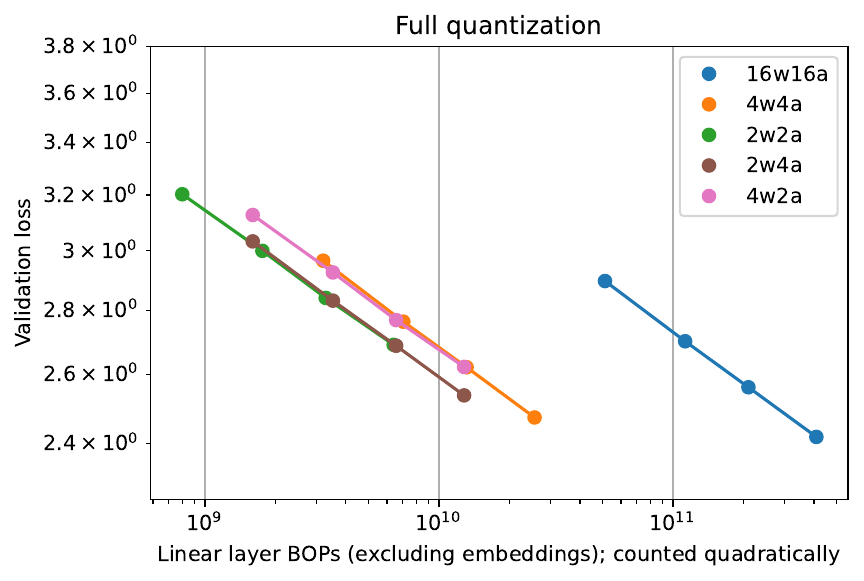}
\caption{Scaling behavior of mixed quantization.}
\label{fig:mixed}
\end{figure}

The final experiment in this series is exploring mixed weight and activation bitwidth. 
The results for various combinations of weight and activation bitwidths are shown in Figure~\ref{fig:mixed}. 

While the overall scaling behavior seems to be as expected, we make the  observation that the Pareto lines of 2W4A and 2W2A as well as of 4W2A and 4W4A essentially overlap. A possible explanation for this effect could be that, unlike weights, activation bitwidth (below some level) scales linearly with the loss. That is, \emph{reducing activation bitwidth by a factor of 2 produces a loss roughly equivalent to a model with half the size}. This could be verified further, for instance, by running experiments with only quantized activations and full precision weights.

\section{Sparsity vs. Quantization}

While scaling behaviour of weight-sparsity is studied in depth in previous work~\citep{frantar2023scalinglawssparselyconnectedfoundation}, we revisit it here under a different setup:  for decoder-only architectures, larger models, and better tuned training parameters. Our goal is to compare the ``effective parameters'' of sparse vs. quantized representations, under similar computational cost. 

First, we find that standard STE-based optimization (top-k sparsification before every forward) works well. This is useful since sparsifying this way involves less hyper-parameter tuning (e.g., gradual pruning schedules) and is hence what adopt this setup across all subsequent experiments.

\paragraph{Method tuning.} However, compared to quantization, applying top-k on entire tensors before every forward pass can be relatively slow; hence, we investigated the importance of global per-tensor top-k in some isolated experiments. 
 Large N:M sparsity with larger values of N and M seems to work almost  identically to full per-tensor top-k, but is much faster ($>$ 2x in our particular setup).

\begin{table*}[h]
\centering
\begin{tabular}{|l|c|}
\hline
\textbf{Llama 100M for 10B tokens @ 50\% sparsity} & \textbf{Validation loss} \\
\hline
2:4 & 3.274 \\
64:128 & 3.223 \\
Per-output & 3.223 \\
Per-tensor & 3.224 \\
\hline
\end{tabular}
\caption{Tuning results for weight sparsity}
\end{table*}

Eventually, we use a \emph{per-row} configuration in our scaling sweep in order to make sure we are not restricting higher sparsity levels too much while still getting substantial speedups from smaller top-k windows.

\begin{figure}[h]
\centering
\includegraphics[width=0.45\textwidth]{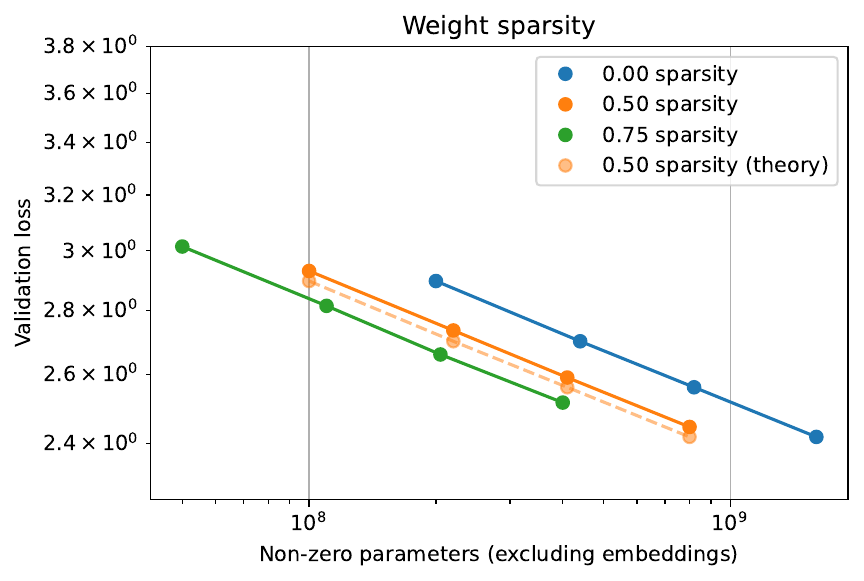}
\includegraphics[width=0.45\textwidth]{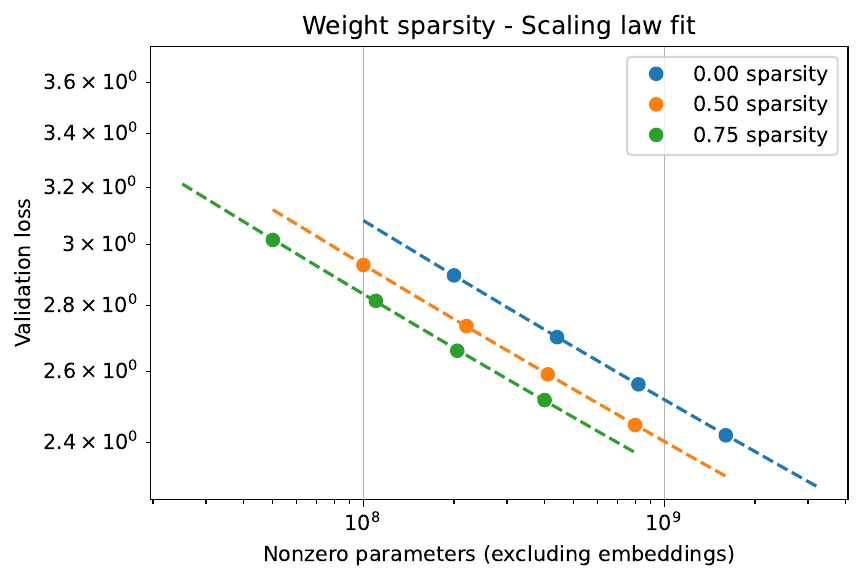}
\caption{Scaling results for weight sparsity.}
\label{fig:sparsity-scaling}
\end{figure}

\paragraph{Scaling.} 
The scaling results for weight sparsity are shown in Figure~\ref{fig:sparsity-scaling}, across three different sparsities, showing a very good law fit. 
As expected, weight sparsity also leads to clear constant multiplier scaling. The size-gain multipliers are approximately 10\% better than identified in  previous work~\citep{frantar2023scalinglawssparselyconnectedfoundation}; this is a reasonable improvement (presumably due to a significantly better optimized setup overall), but still suggests that those multipliers remain relatively stable across many setup changes.

\paragraph{Comparing sparse vs. quantized representations.} 
The multipliers for sparsity are overall significantly lower than for weight-only quantization. However, sparsity also reduces FLOPs proportionally:  remarkably, 50\% sparsity yields a very similar multiplier as 8W8A (0.871 vs 0.857, respectively), which has exactly the same  FLOP reduction under linear scaling. 
Beyond that point, e.g. when comparing full 4-bit quantization (0.747) vs 75\% sparsity (0.622), \emph{quantization  seems to have noticeably better effective parameter count, even under a stringent linear FLOP scaling}. But we note that this balance is largely dependent on the baseline precision and we expect sparsity to become more competitive as we reduce the default precision for our models.  

\begin{table}[h]
\centering
\begin{tabular}{|l|c|c|}
\hline
& 50\% & 75\% \\
\hline
EPM & 0.871 & 0.622 \\
Size-gain & 1.74x & 2.49x \\
\hline
\end{tabular}
\caption{Effectiveness multipliers and size gains for weight sparsity.}
\label{tab:tuning-weight-sparsity}
\end{table}

\section{Related Work}
\label{sec:related-work}

Our work builds on and connects several lines of research around scaling laws, model compression techniques, and the intersection between them.

\paragraph{Scaling Laws for Language Models.} The foundation of this work builds on established scaling laws for language models that characterize how performance improves with model size and training data. \citet{kaplan2020scalinglawsneurallanguage} established the first comprehensive scaling laws showing that loss follows power law relationships with both parameters and data. \citet{hoffmann2022trainingcomputeoptimallargelanguage} refined these results with the Chinchilla scaling laws, suggesting that previous models were over-parameterized and that parameters and data should be scaled roughly equally. Recent work has revealed additional nuances in scaling behavior - for example, when considering data redundancy~\citep{muennighoff2023scaling}, or different model architectures~\citep{clark2022unified}.

\paragraph{Model Compression and Sparsity.} Parallel work has focused on making models more efficient through compression techniques. For sparsity, \citet{frantar2023scalinglawssparselyconnectedfoundation} established the first scaling laws characterizing how sparsity interacts with model and data scaling, showing that sparsity acts as a consistent multiplier on effective parameter count. Their work demonstrated that optimal sparsity levels increase with longer training, as dense models hit diminishing returns. This report directly builds on this earlier work, studying how different representations affect scaling. 

\paragraph{Quantization for Language Models.} 
Recent advances in quantization have enabled dramatically reduced precision while maintaining performance. Post-training quantization methods like GPTQ~\citep{frantar2022gptq} and AWQ~\citep{lin2023awq} have shown strong results for inference. For quantization-aware training, BitNet~\citep{wang2023bitnet} and  follow-up work~\citep{ma2024era1bitllmslarge, kaushal2024spectra} demonstrated stable training with binary and ternary weights, although a precise comparison against dense model scaling is not possible in their setting given the different hyper-parameters used. 

This work complements these efforts by characterizing how quantization during training affects fundamental scaling behavior - showing for instance that weight-only quantization maintains strong parameter efficiency even at very low bitwidths, for both weights and activations. 
In this respect, the thrust of our work is similar to that of concurrent work by~\citet{kumar2024scaling}: relative to their results, we propose a relatively simpler scaling law formulation, albeit in a narrower setting. Moreover, we reveal much more stable precision scaling, since we obtain results suggesting that 4-bit weights and activations may be Pareto-optimal, relative to their findings claiming that there is a precision barrier at around 8-bit precision. In addition, the main goal of our work is different, as we wish to consider a scaling comparison between sparsity and quantization. 

Our work advances this literature by providing the first unified scaling framework encompassing both sparsity and quantization, enabling principled comparison of these compression approaches in the context of large-scale training. The effective parameter framework we propose helps clarify when and how different compression techniques are most beneficial.

\section{Discussion}
\label{sec:discussion}

We have proposed and given evidence of a unified ``effective parameter count'' metric in compresed training of LLMs, revealing a few insights about compression scaling in language models. For instance, through this lens, we find that weight-only quantization maintains strong parameter efficiency even at very low bitwidths, with 4-bit to 1-bit showing approximately 2x improvement. However, when quantizing both weights and activations, we observe clear diminishing returns below 4 bits. This suggests that 4-bit quantization may be optimal for full model quantization. This challenges previous work suggesting ever-lower bitwidths are always better~\citep{ma2024era1bitllmslarge} or that 8-bit presents a clear complexity barrier~\citep{kumar2024scaling}.

One implication concerns the comparison between sparsity and quantization under similar compute constraints. Interestingly, at 50\% sparsity, the effective parameter multiplier (EPM) of 0.871 nearly matches 8-bit quantization (0.857). However, at higher compression rates (e.g., 75\% sparsity vs 4-bit quantization), quantization demonstrates better parameter efficiency. These results suggest that different compression approaches may be optimal depending on the training regime – weight-only quantization provides strong benefits for inference scenarios, while full quantization may be preferable for training efficiency but should likely not go below 4 bits, at least when using the methods from our study.
Importantly, the effectiveness of compression appears to increase with longer training, suggesting particular benefits for heavily-overtrained models. This insight, combined with our unified scaling framework, enables principled selection of compression methods based on compute budgets and deployment scenarios.

Several important questions remain open for future work. First, while our results focus on autoregressive language models, understanding how these scaling relationships transfer to other architectures would be valuable. Second, the potential benefits of combining different compression approaches – for instance, using both sparsity and quantization – remains unexplored. Finally, investigating the role of different quantization schemes (e.g., floating point vs integer) could reveal additional insights into compression scaling.

\bibliographystyle{abbrvnat}
\nobibliography*
\bibliography{main}


\end{document}